\pdfoutput=1

\documentclass[11pt]{article}

\usepackage{acl}

\usepackage{times}
\usepackage{latexsym}

\usepackage[T1]{fontenc}

\usepackage[utf8]{inputenc}

\usepackage{microtype}

\usepackage{graphicx} 
\usepackage{booktabs} 
\usepackage{amsmath} 
\usepackage{multirow}
\usepackage{amssymb}
\usepackage{subfigure}
\usepackage{caption}
\usepackage{algorithm}
\usepackage{algorithmic}
\title{Consistent Representation Learning for Continual Relation Extraction}

\author{Kang Zhao$^{12}$, Hua Xu$^{1*}$, Jiangong Yang$^{12}$, Kai Gao$^{2*}$\\
  $^{1}$State Key Laboratory of Intelligent Technology and Systems, \\Department of Computer Science and Technology, Tsinghua University, Beijing 100084, China \\
  $^{2}$School of Information Science and Engineering, \\Hebei University of Science and Technology, Shijiazhuang 050018, China \\
  {\tt zhaok7878@gmail.com,}\\ 
  {\tt xuhua@tsinghua.edu.cn,yjg219@163.com,gaokai@hebust.edu.cn}\\
  }

\begin{document}
\maketitle
\begin{abstract}
Continual relation extraction (CRE) aims to continuously train a model on data with new relations while avoiding forgetting old ones.
Some previous work has proved that storing a few typical samples of old relations and replaying them when learning new relations can effectively avoid forgetting. However, these memory-based methods tend to overfit the memory samples and perform poorly on imbalanced datasets.
To solve these challenges, a consistent representation learning method is proposed, which maintains the stability of the relation embedding by adopting contrastive learning and knowledge distillation when replaying memory. Specifically, supervised contrastive learning based on a memory bank is first used to train each new task so that the model can effectively learn the relation representation. Then, contrastive replay is conducted of the samples in memory and makes the model retain the knowledge of historical relations through memory knowledge distillation to prevent the catastrophic forgetting of the old task.
The proposed method can better learn consistent representations to alleviate forgetting effectively. Extensive experiments on FewRel and TACRED datasets show that our method significantly outperforms state-of-the-art baselines and yield strong robustness on the imbalanced dataset.
The code is publicly available at \url{https://github.com/thuiar/CRL}.

\end{abstract}

\section{Introduction}
\let\thefootnote\relax\footnotetext{*Corresponding Author}
Relation extraction (RE) is an essential issue in information
extraction (IE), which can apply to many downstream NLP tasks, such as information retrieval \cite{DBLP:conf/www/XiongPC17} and question and answer \cite{DBLP:conf/ijcai/TaoGSWZY18}. 
For example, given a sentence $x$ with the annotated entities pairs $e_{1}$ and $e_{2}$, the RE aims to identify the relations between $e_{1}$ and $e_{2}$. However, traditional relation extraction models \cite{zhou2016attention,soares2019matching} always assume a fixed set of predefined relations and train on a fixed dataset, which cannot handle the growing relation types in real life well.

To solve this situation, continual relation extraction (CRE) is introduced \cite{wang2019sentence, han2020continual,wu2021curriculum,cui2021refining}. Compared with traditional relation extraction, CRE aims to help the model learn new relations while maintaining accurate classification of old ones. \citet{wang2019sentence} shows that continual relation learning needs to alleviate the catastrophic forgetting of old tasks when the model learns new tasks. 
Because neural networks need to retrain a fixed set of parameters with each training, the most efficient solution to the problem of catastrophic forgetting is to store all the historical data and retrain the model with all the data each time a new relational instance appears.
This method can achieve the best effect in continual relation learning, but it is not adopted in real life due to the time and computing power costs.

Some recent works have proposed a variety of methods to alleviate the catastrophic forgetting problem in continual learning, including regularization methods \cite{kirkpatrick2017overcoming, zenke2017continual, liu2018rotate}, dynamic architecture methods \cite{chen2015net2net, fernando2017pathnet}, and memory-based methods \cite{lopez2017gradient, chaudhry2018efficient}. Although these methods have been verified in simple image classification tasks, previous works have proved that memory-based methods are the most effective in natural language processing applications \cite{wang2019sentence, d2019episodic}. In recent years, the memory-based continual relation extraction model has made significant progress in alleviating the problem of catastrophic forgetting \cite{han2020continual, wu2021curriculum, cui2021refining}. 
\citet{wang2019sentence} proposes a mechanism for embedding sentence alignment in memory maintenance to ensure the stability of the embedding space.
\citet{han2020continual} introduces a multi-round joint training process for memory consolidation. But these two methods only explore the problem of catastrophic forgetting in the overall performance of the task sequence. 
\citet{wu2021curriculum} proposes to integrate curriculum learning. Although it is possible to analyze the characteristics of each subtask and the performance of the corresponding model, it still fails to make full use of the saved sample information.
\citet{cui2021refining} introduce an attention network to refine the prototype to better recover the interruption of the embedded space.
However, this method will produce a bias in the classification of the old task as the new task continues to learn the classifier, which will affect the performance of the old task.
Although the above method can alleviate catastrophic forgetting to a certain extent, it does not consider the consistency of relation embedding space.

Because the performance of the model of CRE is sensitive to the quality of sample embedding, it needs to ensure that the learning of new tasks will not damage the embedding of old tasks.
Inspired by supervised contrastive Learning \cite{khosla2020supervised} to explicitly constrain data embeddings, a consistent representation learning method is proposed for continual relation extraction, which constrains the embedding of old tasks not to occur significantly change through supervised contrastive learning and knowledge distillation.
Specifically, the example encoder first trains on the current task data through supervised contrastive learning based on memory bank, and then uses k-means to select representative samples to storage as memory after the training is completed.
To relieve catastrophic forgetting, contrastive replay is used to train memorized samples. At the same time, to ensure that the embedding of historical relations does not undergo significant changes, knowledge distillation is used to make the embedding distribution of the new and old tasks consistent. In the testing phase, the nearest class mean (NCM) classifier is used to classify the test sample, which will not be affected by the deviation of the classifier.

In summary, our contributions in this paper are summarized as follows:
First, a novel CRE method is proposed, which uses supervised contrastive learning and knowledge distillation to learn consistent relation representations for continual learning.
Second, consistent representation learning can ensure the stability of the relational embedding space to alleviate catastrophic forgetting and make full use of stored samples.
Finally, extensive experiments results on FewRel and TACRED datasets show that the proposed method is better than the latest baseline and effectively mitigates catastrophic forgetting.

\section{Related Work}

\subsection{Continual Learning}
Existing continual learning models mainly focus on three areas: (1) Regularization-based methods \cite{kirkpatrick2017overcoming,zenke2017continual} impose constraints on updating neural weights important to previous tasks for relieving catastrophic forgetting. (2) Dynamic architecture methods \cite{chen2015net2net,fernando2017pathnet} extends the model architecture dynamically to learn new tasks and prevent forgetting old tasks effectively. However, these methods are unsuitable for NLP applications because the model size increases dramatically with increasing tasks. (3) Memory-based methods \cite{lopez2017gradient, aljundi2018memory, chaudhry2018efficient,mai2021supervised} saves some samples from old tasks and continuously learns them in new tasks to alleviate catastrophic forgetting.
\citet{dong2021few} proposes a simple relational distillation incremental learning framework to balance retaining old knowledge and adapting to new knowledge.
\citet{yan2021dynamically} proposes a new two-stage learning method that uses dynamic expandable representation for more effective incremental conceptual modelling.
Among these methods, memory-based methods are the most effective in NLP tasks \cite{wang2019sentence,sun2019lamol,d2019episodic}.
Inspired by the success of memory-based methods in the field of NLP, we use the framework of memory replay to learn new relations that are constantly emerging.

\subsection{Contrastive Learning}
Contrastive learning (CL) aims to make the representations of similar samples map closer to each other in the embedded space, while that of dissimilar samples should be farther away \cite{jaiswal2021survey}. In recent years, the rise of CL has made great progress in self-supervised representation learning. \cite{wu2018unsupervised,he2020momentum,li2020prototypical,chen2021exploring}.
The common point of these works is that no labels are available, so positive and negative pairs were formed through data augmentations.
Recently, supervised contrastive learning \cite{khosla2020supervised} has received much attention, which uses label information to extend contrastive learning.
\citet{hendrycks2019benchmarking} compares the supervised contrastive loss with the cross-entropy loss on the ImageNet-C dataset, and verifies that the supervised contrastive loss is not sensitive to the hyperparameter settings of the optimizer or data enhancement.
\citet{chen2020simple} proposed a contrastive learning framework for visual representations that does not require a special architecture or memory bank. 
\citet{khosla2020supervised} extend the self-supervised batch contrastive approach to the fully-supervised setting, which use supervised contrastive loss learning better represetation. 
\citet{liu2020hybrid} proposed a hybrid discriminant-generative training method based on an energy model. 
In this paper, contrastive learning is applied to continual relation extraction to extract better relation representation.

\section{Methodology}
\subsection{Problem Formulation}

In continual relation extraction, given a series of $K$ tasks $\{T_{1},T_{2},...,T_{K}\}$, where the k-th task has its own training set $D_{k}$ and relation set $R_{k}$.
Each task $T_{k}$ is a traditional supervised classification task, including a series of examples and their corresponding labels $\{(x_{i}, y_{i})\}_{i=1}^{N}$, where $x_{i}$ is the input data, including the natural language text and entity pair, and $y_{i}\in R_{k}$ is the relation label. 
The goal of continual relation learning is to train the model, which keeps learning new tasks while avoiding catastrophic forgetting of previous learning tasks.
In other words, after learning the $k$-th task, the model can identify the relation of a given entity pair into $\hat{R}_{k}$, where $\hat{R}_{k}=\cup_{i=1}^{k}R_{i}$ is the relation set already observed till the $k$-th task.

In order to mitigate catastrophic forgetting in continual relational extraction, episodic memory modules have been used in previous work \cite{wang2019sentence, han2020continual, cui2021refining},  to store small samples in historical tasks. 
Inspired by \cite{cui2021refining}, we store several representative samples for each relation. Therefore, the episodic memory module for the observed relations in $T_{1}\sim T_{k}$ is $\hat{M}_{k}=\cup_{r\in\hat{R}_{k}}M_{r}$, where $M_{r}=\left\{ (x_{i},y_{i})\right\}_{i=1}^{O}$, $r$ represents a certain relation, and $O$ is sample number (memory size).

\begin{algorithm}[H]
\caption{ Training procedure for $T_{k}$} 
\label{alg:Framwork} 
\begin{algorithmic}[1] 
\REQUIRE ~~\\ 
The training set of $D_{k}$ of the $k$-th task, encoder $\mathbf{E}$, projection head $\mathrm{Proj}$, history memory $M_{k-1}$, current relation set $R_{k}$, history relation set $\hat{R}_{k-1}$\\
\ENSURE ~~\\ 
encoder $f_{k}(\cdot)$, history memory $M_{k}$, history relation set $\hat{R}_{k}$ \\
\IF{$T_{k}$ is not the first task}
\STATE get memory knowledge with $\mathbf{E}$ on $M_{k-1}$;
\ENDIF
\STATE $M_{b} \gets \mathbf{E}(D_{k})$ ; 
\label{ code:fram:extract }
\FOR{$i \gets 1$ to $epoch_{1}$}
\FOR{ $each$ $x_{j} \in D_{k}$}
\STATE Sample from $M_{b}$;
\STATE Update $\mathbf{E}$ and $\mathrm{Proj}$ with $\nabla L_{CL}$;
\STATE Update $M_{b}$;
\ENDFOR
\ENDFOR
\STATE Select informative examples from $D_{k}$ to store into $\hat{M}$
\STATE $M_{k} \gets M_{k-1} \cup \hat{M}$;
\STATE $\hat{R}_{k} \gets \hat{R}_{k-1} \cup R_{k}$;
\IF{$T_{k}$ is not the first task}
\STATE $\tilde{M_{b}} \gets \mathbf{E}(M_{k})$ ;
\FOR{$i \gets 1$ to $epoch_{2}$}
\FOR{$each$ $x_{j} \in M_{k}$}
\STATE Sample from $\tilde{M_{b}}$;
\STATE Update $\mathbf{E}$ and $\mathrm{Proj}$ with $\nabla L_{CR}$ and $\nabla L_{KL}$;
\STATE Update $\tilde{M_{b}}$;
\ENDFOR
\ENDFOR
\STATE Select informative examples from $D_{k}$ to store into $\hat{M}$;
\STATE $M_{k} \gets M_{k-1} \cup \hat{M}$
\ENDIF
\RETURN $\mathbf{E}$, $M_{k}$, $\hat{R}_{k}$; 
\end{algorithmic}
\end{algorithm}

\begin{figure*}[t]
	\centering
	\scalebox{0.85}{
		\includegraphics{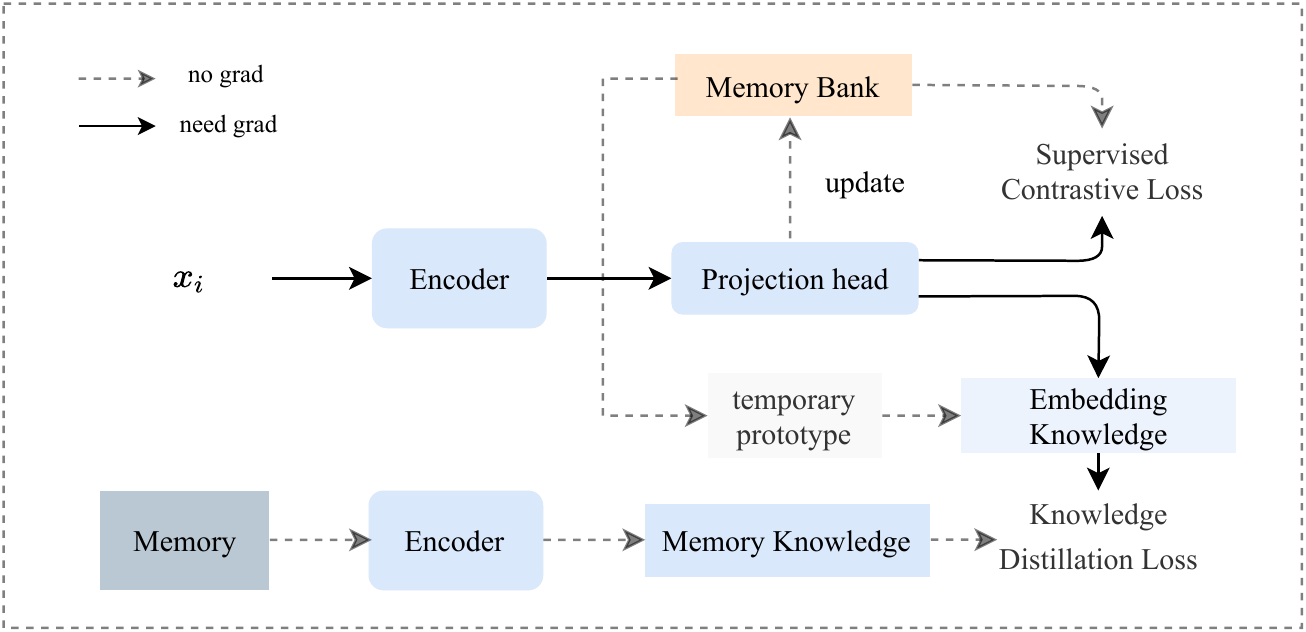}} 
	\caption{Framwork of consistent representation learning.}
	\label{fig1}
\end{figure*}

\subsection{Framework}

The consistent representation learning (CRL) in the current task is described in Algorithm \ref{alg:Framwork}, which consists of three main steps:
(1) \textbf{Init training for new task} (line $4\sim 11$): The parameters of the encoder and projector head are trained on the training sample in $D_{k}$ with supervised contrastive learning.
(2) \textbf{Sample selection} (line $12\sim 13$): For each relation $r\in R_{k}$, we retrieve all samples labeled $r$ from $D_{k}$. Then, the k-means algorithm is used to cluster the samples. The relation representation of the sample closest to the center is selected and stored in memory for each cluster.
(3) \textbf{Consistent representation learning} ($16\sim 23$): In order to keep the embedding of historical relations in space consistent after learning new tasks, we perform contrastive replay and knowledge distillation constraints on the samples in memory. 

\subsection{Encoder}
The key of CRE is to obtain a better relation representation. The pre-trained language model BERT \cite{DBLP:conf/naacl/DevlinCLT19} shows a powerful ability in extracting contextual representation of text. Therefore, BERT is used to encode entity pairs and context information to get the relational representation. 

Given a sentence $x=[w_{1}, \dots, w_{|x|}]$ and a pair of entities $(\mathrm{E1}, \mathrm{E2})$, we follow \citet{DBLP:conf/acl/SoaresFLK19} augment $x$ with four reserved word pieces to mark the begin and end of each entity mentioned in the sentence.
The new token sequence is fed into BERT instead of $x$.
To get the final relation representation between the two entities, the output corresponding to the positions of $\mathrm{E1}$ and $\mathrm{E2}$ are concatenated, and then map it to a high-dimensional hidden representation $\mathbf{h} \in \mathbb{R}^{d_{h}}$, as follows:
\begin{equation}
\begin{aligned}
\mathbf{h}  =& \mathbf{W} [ \mathbf{h}_{[\mathrm{E1}]};\mathbf{h}_{[\mathrm{E2}]}]+\mathbf{b},
\end{aligned}
\label{111}
\end{equation}
where $\mathbf{W} \in \mathbb{R}^{2d_{h} \times d_{h} } $ and $\mathbf{b} \in \mathbb{R}^{d_{h}}$ are trainable parameters.
The encoder in which the above-mentioned encoded sentence is a relation representation is denoted as $\mathbf{E}$.

Then, we use a projection head $\mathrm{Proj}$ to obtain the low-dimensional embedding:
\begin{equation}
\begin{aligned}
\mathbf{\tilde{z}}  =& \mathrm{Proj}(\mathbf{h}),
\end{aligned}
\label{111}
\end{equation}
where $\mathrm{Proj}(\cdot)=\mathrm{MLP}(\cdot)$ is composed of two layers of neural networks.
The normalized embedding $\mathbf{z}=\tilde{\mathbf{z}}/||\tilde{\mathbf{z}}||$ is used for contrastive learning, and the hidden representation is used for classification.

\subsection{Inital training for new task}
\label{sec3.4}
Before training for each new task $T_{k}$, we first use Encoder to extract the embedding $\tilde{\mathbf{z}}$ of the relational representation of each sentence in $D_{k}$, and use them as the initialized memory bank $M_{b}$:
\begin{equation}
\begin{aligned}
M_{b} \gets \{\mathbf{z}_{i}\}_{i=1}^{N}.
\end{aligned}
\label{111}
\end{equation}

At the beginning of training, relation representation extraction is performed on each batch $B$. Then the data embedding is explicitly constrained by clustering through supervised contrastive learning \cite{khosla2020supervised}:
\begin{equation}
\begin{aligned}
\mathcal{L}_{\mathrm{CL}}
=\sum_{i \in I} \frac{-1}{|P(i)|} \sum_{p \in P(i)} \log \frac{\exp \left(z_{i} \cdot z_{p} / \tau\right)}{\sum_{j \in S_{I}} \exp \left(z_{i} \cdot z_{j} / \tau\right)},
\end{aligned}
\label{scl}
\end{equation}
where $I=\{1,2,\dots, |B|\}$ is the set of indices of $B$. $S_{I}$ represents the indices set of randomly sampled partial samples from $M_{b}$. $P(i)=\{p\in S_{I}: y_{p}=y_{i} \}$ is the indices set that is the same as the $z_{i}$ label in $M_{b}$, and $|P(i)|$ is its cardinality. $\tau \in \mathcal{R}^{+}$ is an adjustable temperature parameter controling the separation of classes, the $\cdot$ indicates the dot product.

After backpropagating the gradient of loss on each batch, we update the representation in the memory bank:
\begin{equation}
\begin{aligned}
M_{b}[\tilde{I}] \gets \{\mathbf{z}_{i}\}_{i=1}^{|B|}.
\end{aligned}
\label{111}
\end{equation}
where $\tilde{I}$ is the corresponding index set of this batch of samples in $M_{b}$.
After $epoch1$ training set training, the model can learn a better relation representation.

\subsection{Selecting Typical Samples for Memory}
In order to make the model not forget the relevant knowledge of the old task when it learns the new task, some samples need to be stored in $M_{r}$. Inspired by \cite{han2020continual,cui2021refining}, we use k-means to cluster each relation, where the number of clusters is the number of samples that need to be stored for each class. Then, the relation representation closest to the center is selected and stored in memory for each cluster.

\subsection{Consistent Representation Learning}
After learning a new task, the representation of the old relation in the space may change. In order to make the encoder not change the knowledge of the old task while learning the new task, we propose two replay strategies to learn consistent representation for alleviating this problem: contrastive replay and knowledge distillation. Figure \ref{fig1} shows the main flow of consistent representation learning.

\paragraph{Contrastive Replay with Memory Bank}
After the new task learning is over, we use the new task to train the encoder to further train the encoder by replaying the samples stored in memory $M_{k}$.
After the learning of the current task is over, we use the same method in Section \ref{sec3.4} to replay the samples stored in memory $M_{k}$.

The difference here is that each batch uses all the samples in the entire memory bank for contrastive learning, as follows:
\begin{equation}
\begin{aligned}
\mathcal{L}_{\mathrm{CR}}
=\sum_{i \in I} \frac{-1}{|P(i)|} \sum_{p \in P(i)} \log \frac{\exp \left(z_{i} \cdot z_{p} / \tau\right)}{\sum_{j \in \tilde{S}_{I}} \exp \left(z_{i} \cdot z_{j} / \tau\right)},
\end{aligned}
\label{111}
\end{equation}
where $\tilde{S}_{I}$ represents the set of indices of all samples in $\tilde{M}_{b}$. $\tilde{M}_{b}$ is the memory bank, which stores the normalized representation of all samples in $M_{k}$.

By replaying the samples in memory, the encoder can alleviate the forgetting of previously learned knowledge, and at the same time, consolidate the knowledge learned in the current task. However, contrastive replay allows the encoder to train on a small number of samples, which risks overfitting. On the other hand, it may change the distribution of relations in the previous task. Therefore, we propose knowledge distillation to make up for this shortcoming.

\paragraph{Knowledge Distillation for Relieve Forgetting}
We hope that the model can retain the semantic knowledge between relations in historical tasks. Therefore, before the encoder is trained on a task, we use the similarity metric between the relations in memory as \textbf{Memory Knowledge}. Then use the knowledge distillation to relieve the model from forgetting this knowledge.

Specifically, the samples in the memory are encoded first, and then the prototype of each class is calculated:
\begin{equation}
\begin{aligned}
p_{c}=\sum_{i=1}^{O}z_{i}^{c},
\end{aligned}
\end{equation}
where $O$ is the number of memory size, $z_{i}^{c}$ is the relation representation belonging to class $c$.
Then, the cosine similarity between the classes is calculated to represent the knowledge learned in the memory:
\begin{equation}
\begin{aligned}
a_{ij}=\frac{p_{i}^{T}p_{j}}{\left \| p_{i} \right \| \left \| p_{j} \right \|},
\end{aligned}
\end{equation}
where $a_{ij}$ is the cosine similarity between prototype $i$ and $j$.

When performing memory replay, we use KL divergence to make the encoder retain the knowledge of the old task.
\begin{equation}
\begin{aligned}
\mathcal{L}_{KL}=\textstyle \sum_{i} KL(P_{i}||Q_{i}),
\end{aligned}
\end{equation}
where $P_{i} = \{p_{ij}\}_{j=1}^{|\hat{R}_{k}|}$ is the metric distribution of the prototype before training, and $p_{ij}=\frac{\exp \left(a_{ij}/\tau\right)}{\sum_{j} \exp \left(a_{ij} / \tau\right)}$. Similarly, $Q_{i} = \{q_{ij}\}_{j=1}^{|\hat{R}_{k}|}$ is the metric distribution of calculate the temporary prototype from the memory bank during training, and $q_{ij} = \frac{\exp \left(\tilde{a} _{ij}/\tau\right)}{\sum_{j} \exp \left(\tilde{a}_{ij} / \tau\right)}$. 
$\tilde{a}$ is the \textbf{Embedding Knowledge} of the memory $M_{k}$, which is the cosine similarity between temporary prototypes.
The temporary prototype is dynamically calculated in each batch based on the memory bank $\tilde{M_{b}}$.

\begin{table*}[ht]
\centering
\small
\begin{tabular}{lcccccccccc}
\toprule
\multicolumn{11}{c}{FewRel}                                                                                                                                                                                                                                                                                            \\ \hline
\addlinespace[0.1cm]
\multicolumn{1}{l|}{Model}     & T1                        & T2                        & T3                        & T4                        & T5                        & T6                        & T7                        & T8                        & T9                        & T10                       \\ \hline \addlinespace[0.1cm]
\multicolumn{1}{l|}{EA-EMR}    & 89.0                        & 69.0                        & 59.1                      & 54.2                      & 47.8                      & 46.1                      & 43.1                      & 40.7                      & 38.6                      & 35.2                      \\
\multicolumn{1}{l|}{EMAR}      & 88.5                      & 73.2                      & 66.6                      & 63.8                      & 55.8                      & 54.3                      & 52.9                      & 50.9                      & 48.8                      & 46.3                      \\
\multicolumn{1}{l|}{CML}       & 91.2                      & 74.8                      & 68.2                      & 58.2                      & 53.7                      & 50.4                      & 47.8                      & 44.4                      & 43.1                      & 39.7                      \\
\multicolumn{1}{l|}{EMAR+BERT} & \textbf{98.8}                      & 89.1                      & 89.5                      & 85.7                      & 83.6                      & 84.8                      & 79.3                      & 80.0                        & 77.1                      & 73.8                      \\
\multicolumn{1}{l|}{RP-CRE}    & 97.9                      & 92.7                      & 91.6                      & 89.2                      & 88.4                      & 86.8                      & 85.1                      & 84.1                      & 82.2                      & 81.5                      \\
\multicolumn{1}{l|}{RP-CRE\dag}   & 97.8 & \textbf{95.1} & 91.8 & 90.5 & \textbf{89.9} & 87.7 & 86.6 & 85.6 & 84.3 & 82.6 \\ \hline \addlinespace[0.1cm]
\multicolumn{1}{l|}{CRL}       & 98.2                      & 94.6                      & \textbf{92.5}                      & \textbf{90.5}                        & 89.4                        & \textbf{87.9}                      & 86.9                      & \textbf{85.6}                      & \textbf{84.5}                      & \textbf{83.1}                      \\ 
\multicolumn{1}{c|}{w/o KL}   & 98.2    & 94.6                      & 92.4                      & 90.5                      & 89.5                      & 87.7                      & \textbf{87.1}                      & 85.4                      & 84.2                      & 82.7                      \\

\multicolumn{1}{c|}{w/o CR}   & 98.2    & 94.7                      & 92.0                      & 90.2                      & 88.9                      & 87.1                      & 85.8                      & 84.6                      & 83.0                      & 81.5                      \\
\hline \addlinespace[0.1cm]
\toprule
\multicolumn{11}{c}{TACRED}                                                                                                                                                                                                                                                                                            \\ \hline \addlinespace[0.1cm]
\multicolumn{1}{l|}{Model}     & T1                        & T2                        & T3                        & T4                        & T5                        & T6                        & T7                        & T8                        & T9                        & T10                       \\ \hline \addlinespace[0.1cm]
\multicolumn{1}{l|}{EA-EMR}    & 47.5                      & 40.1                      & 38.3                      & 29.9                      & 24                        & 27.3                      & 26.9                      & 25.8                      & 22.9                      & 19.8                      \\
\multicolumn{1}{l|}{EMAR}      & 73.6                      & 57.0                        & 48.3                      & 42.3                      & 37.7                      & 34.0
& 32.6                      & 30.0                        & 27.6                      & 25.1                      \\
\multicolumn{1}{l|}{CML}       & 57.2                      & 51.4                      & 41.3                      & 39.3                      & 35.9                      & 28.9                      & 27.3                      & 26.9                      & 24.8                      & 23.4                      \\
\multicolumn{1}{l|}{EMAR+BERT} & 96.6                      & 85.7                      & 81                        & 78.6                      & 73.9                      & 72.3                      & 71.7                      & 72.2                      & 72.6                      & 71.0                        \\
\multicolumn{1}{l|}{RP-CRE}    & 97.6                      & 90.6                      & 86.1                      & 82.4                      & 79.8                      & 77.2                      & 75.1                      & 73.7                      & 72.4                      & 72.4                      \\
\multicolumn{1}{l|}{RP-CRE\dag}   & 97.6                      & 93.1                      & \textbf{90.6}                        & \textbf{85.1}                      & 82.7                      & 81.1                      & 78.3                      & 76.0                      & 76.1                      & 75.7                      \\ \hline \addlinespace[0.1cm]
\multicolumn{1}{l|}{CRL}       & \textbf{97.7}                      & 93.2                      & 89.8                      & 84.7                      & 84.1                      & 81.3                      & \textbf{80.2}                      & 79.1                      & \textbf{79.0}                      & \textbf{78.0}                      \\ 
\multicolumn{1}{c|}{w/o KL}       & 97.7                      & \textbf{94.3}                      & 90.1                      & 84.9                      & \textbf{84.7}                      & \textbf{82.5}                      & 80.0                      & \textbf{79.2}                      & 79.0                      & 77.7                      \\
\multicolumn{1}{c|}{w/o CR}       & 97.7                      & 92.7                      & 88.8                      & 84.7                      & 82.3                      & 80.5                      & 77.8                      & 75.9                      & 75.2                      & 74.3                      \\
\hline
\end{tabular}
\caption{Accuracy (\%) on all observed relations (which will continue to accumlate over time) at the stage of learning current task. The method marked by $\dag$  represents the results generated from open source code$^{1}$ and the other baseline results copied from the original paper \cite{cui2021refining}. All results are compared at memory size = 10.}
\label{major}
\end{table*}

\subsection{NCM for Prediction}

To predict a label for a test sample $x$, the nearest class mean (NCM) \cite{mai2021supervised} compares the embedding of $x$ with all the prototypes of memory and assigns the class label with the most similar prototype:
\begin{equation}
\begin{aligned}
p_{c}=&\frac{1}{n_{c}} \sum_{i} \mathbf{E}\left(\bar{x}_{i}\right) \cdot \mathbb{1}\left\{y_{i}=c\right\}, \\
y^{*}=&\underset{c=1, \ldots, k}{\operatorname{argmin}}\left\|f(\mathrm{x})-p_{c}\right\|,
\end{aligned}
\label{111}
\end{equation}
where $\bar{x}\in M_{k}$ is stored sample, and $y^{*}$ is a predicted label. Since the NCM classifier compares the embedding of the test sample with prototypes, it does not require an additional FC layer. Therefore, new classes can be added without any architecture modification.

\section{Experiments}

\subsection{Datasets}
Our experiments are conducted on two benchmark datasets: in the experiment, the training-test-validation that the split ratio is 3:1:1.

\let\thefootnote\relax\footnotetext{$^{1}$https://github.com/fd2014cl/RP-CRE}
\paragraph{FewRel} \cite{han-etal-2018-fewrel} It is a RE dataset that contains 80 relations, each with 700 instances. Following the experimental settings by \citet{wang2019sentence}, the original train and valid set of FewRel are used for experimental, which contains 80 classes. 

\paragraph{TACRED} \cite{zhang2017position} It is a large-scale RE dataset containing 42 relations (including no relations) and 106,264 samples, built on news networks and online documents.
Compared with FewRel, the samples in TACRED are imbalanced. Following \citet{cui2021refining}, the number of training samples for each relation is limited to 320 and the number of test samples of relation to 40.

\subsection{Evaluation Metrics}
Average accuracy is a better measure of the effect of catastrophic forgetting because it emphasizes the model's performance on earlier tasks \cite{han2020continual, cui2021refining}. This paper evaluates the model by using the average accuracy of $K$ tasks at each step.

\subsection{Baselines}
We evaluate CRL and several baselines on benchmarks for comparison: 

(1) EA-EMR \cite{wang2019sentence} introduced a memory replay and embedding alignment mechanism to maintain memory and alleviate embedding distortion during training for new tasks.

(2) EMAR \cite{han2020continual} constructs a memory activation and reconsolidation mechanism to alleviate the catastrophic forgetting problem in CRE. 

(3) CML \cite{wu2021curriculum} proposed a curriculum-meta learning method to alleviate the order sensitivity and catastrophic forgetting in CRE. 

(4) RP-CRE \cite{cui2021refining} achieves enhanced performance by utilizing relation prototypes to refine sample embeddings, thereby effectively avoiding catastrophic forgetting.



\begin{figure*}[h]
\subfigure[Results on FewRel.] 
{
	\begin{minipage}[b]{.5\linewidth}
		\centering
		\centerline{\includegraphics[width=7.7cm]{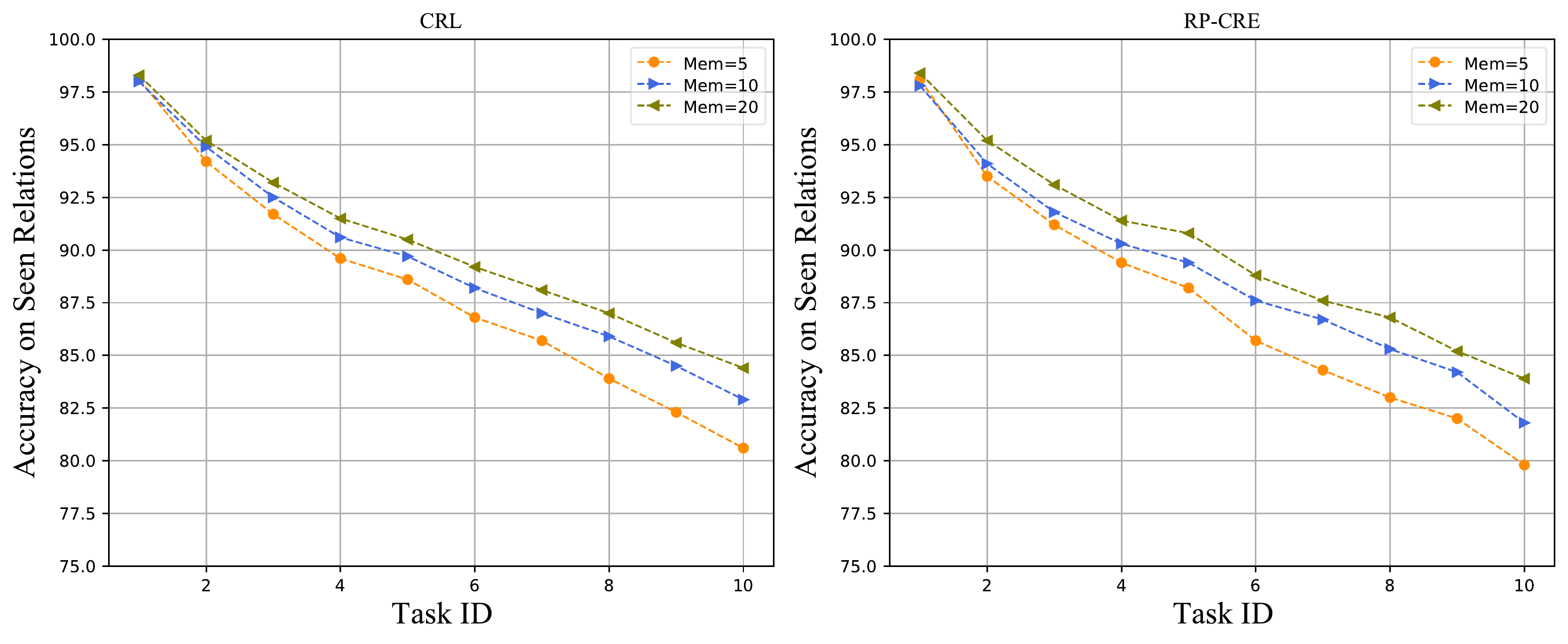}}
	\end{minipage}
}
\hfill
\subfigure[Results on TACRED.] 
{
	\begin{minipage}[b]{0.5\linewidth}
		\centering
		\centerline{\includegraphics[width=7.7cm]{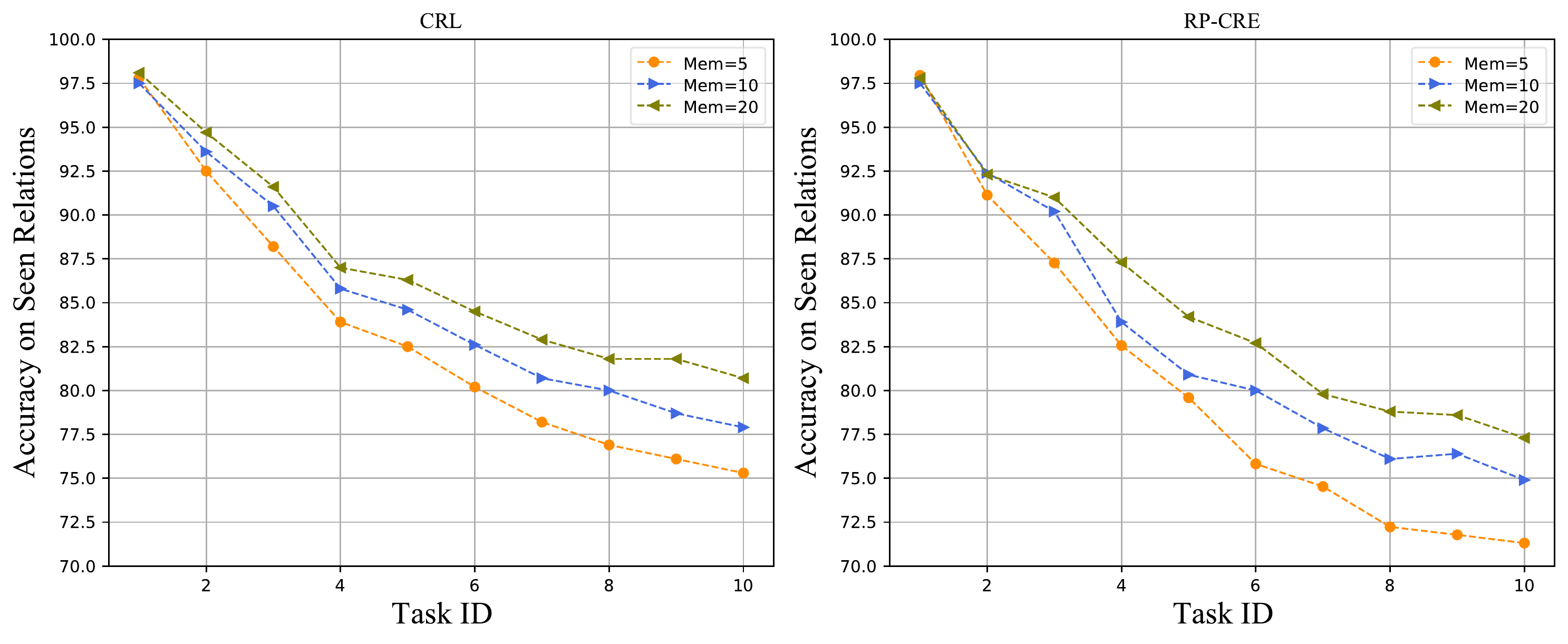}}
	\end{minipage}
}
	\caption{Comparison of model's dependence on memory size, it shows that our model has a light dependence on memory size. The X-axis is the serial ID of the current task, Y-axis is the accuracy of the standard model on the test set from all observed relations at current stage.}
	\label{fig2}
\end{figure*}

\subsection{Training Details and Parameters Setting}
A completely random sampling strategy at the relation level is adopted. It simulates ten tasks by randomly dividing all relations of the dataset into 10 sets to simulate 10 tasks, as suggested in \cite{cui2021refining}. 
For a fair comparison, we set the random seed of the experiment to be the same as the seed in \cite{cui2021refining}, so that the task sequence is exactly the same. Note that our reproduced model RP-CRE $\dag$ and CRL use strictly the same experimental environment. In order to facilitate the reproduction of our experimental results, the proposed method source code and detailed hyperparameters are provided on Github$^{2}$.

\let\thefootnote\relax\footnotetext{$^{2}$\url{https://github.com/thuiar/CRL}}

\subsection{Results and Discussion}
Table \ref{major} shows the results of the proposed methods and baselines ones compared on two datasets, where RP-CRE $\dag$ is reproduced under the same conditions based on open source code. We also ablated knowledge distillation and contrastive replay for consistent representation learning. CRL (w/o KL) and CRL (w/o CR) respectively refer to removing knowledge distillation loss $\mathcal{L}_{KL}$ and contrastive replay loss $\mathcal{L}_{CR}$ when replaying memory.
From the table, some conclusions can be drawn:

(1) Our proposed CRL is significantly better than other baselines and achieves state-of-the-art performance in the vast majority of settings. Compared with RP-CRE, our model also produces apparent advantages. It proves that CRL can learn better consistent relation representations and is more stable in the process of continual learning.

(2) It is observed that all baselines perform worse on the TACRED dataset. The primary reason for this result is that TACRED is an imbalanced dataset. However, our model performs better than RP-CRE's last task on TACRED (3.4\% higher than RP-CRE), which is more significant than the improvement (0.5\%) on the class-balanced dataset FewRel. It shows that our model is more robust to scenarios with class-imbalanced.

(3) Comparing CRL and CRL (w/o KL), not adopting knowledge distillation during training can cause the model to drop 1\% and 0.6\% on FewRel and TACRED, respectively. The experimental results show that knowledge distillation can uniformly alleviate the model's forgetting of previous knowledge to learn a better consistent representation.

(4) Comparing CRL and CRL (w/o CR), removing L during memory replay caused the model to drop 2.4\% and 4.8\% on FewRel and TACRED, respectively. The reason for the significant drop is that only adopting $\mathcal{L}_{KL}$ cannot make the model review the samples of the current task, which leads to overfitting in the historical relations during replay.

\begin{figure*}[t]
	\centering
	\scalebox{0.4}{
		\includegraphics{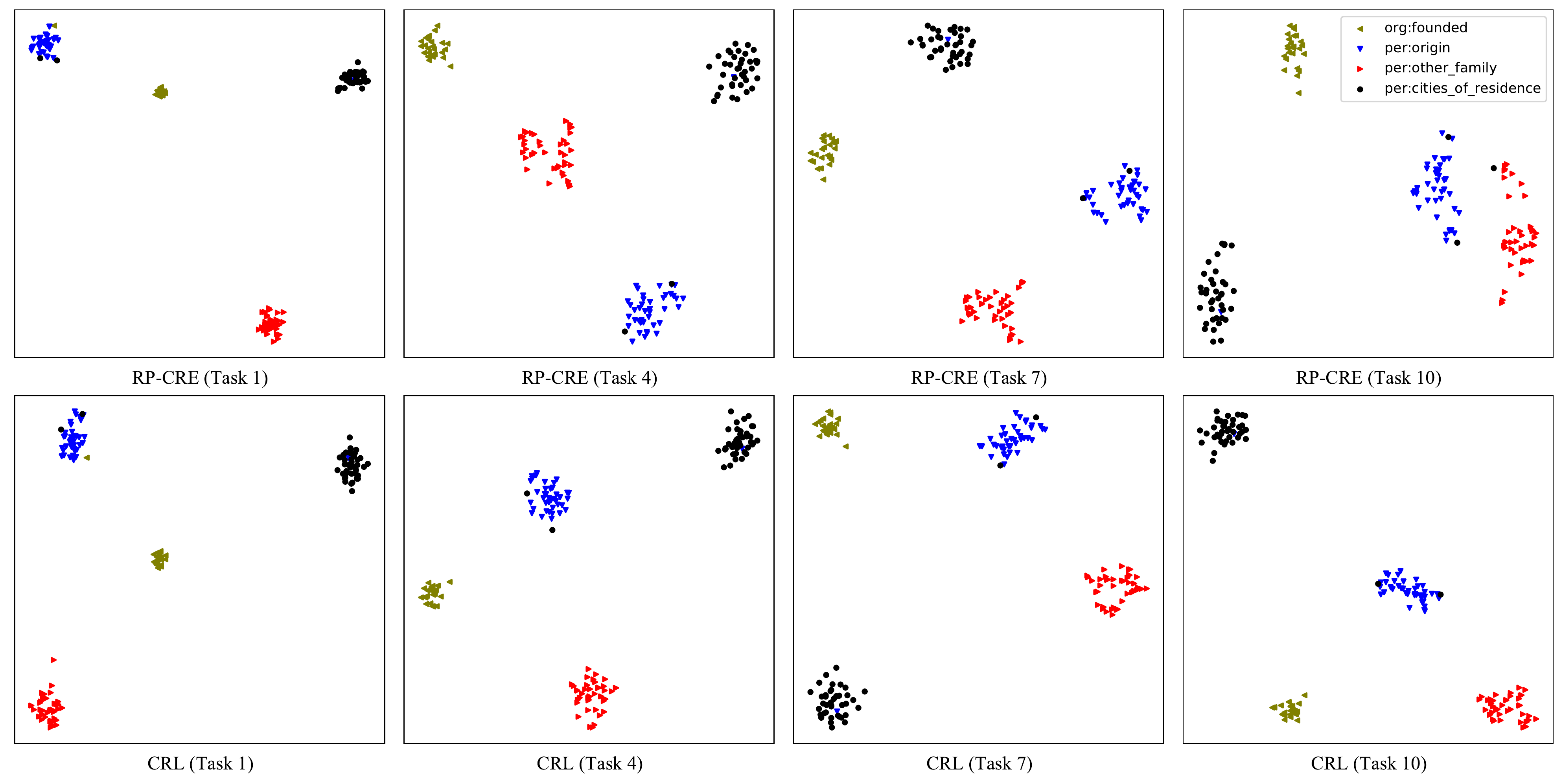}} 
	\caption{A visualization of relation represetation learnted from task 1 test set by RP-CRE and CRL at different task.}
	\label{fig3}
\end{figure*}

\subsection{Effect of Memory Size}
The memory size is the number of memory samples needed for each relation. In this section, we will study the impact of memory size on the performance of our model and RP-CRE. We compare three memory sizes: 5, 10, and 20. The experimental results are shown in Figure \ref{fig2}.

We choose RP-CRE as the main competitor, where all configurations and task sequence remain unchanged. (1) As the size of the memory decreases, the performance of the model tends to decline, which shows that the size of the memory is a key factor that affects continuous learning and learning. But our model is more stable than RP-CRE (the performance gap in the final task), especially on the TACRED dataset.
(2) On both FewRel and TACRED, CRL keeps the best performance under different memory sizes and produces obvious advantages in small memory. It indicates that utilizing consistent representation learning is a more effective way to utilize memory than the existing memory-based CRE method.

\subsection{Effect of Consistent Representation Learning}
In order to explore the long-term effects of consistency representation learning in continual relation extraction, we tested our model and RP-CRE on TACRED to observe the changes in the embedding space of old tasks as new tasks continue to increase. The model performs feature extraction on all samples in the test set in task 1 at the end of tasks 1, 4, 7, and 10. Then t-SNE is used to represent the dimensionality reduction relation representation. All samples on the test set of task 1 are drawn, where different color points represent different ground-truth labels. The visualization results are shown in Figure \ref{fig3}.

From Figure \ref{fig3}, we can see that although the relation embeddings of RP-CRE are clustered and separated in each class after prototype refinement, as new tasks are continuously learned, the data embedding of task 1 is obviously scattered. In contrast, our model retains a good separation between classes, while the data embedding within classes is compact and has a certain diversity. In addition, we can see that our model has relatively stable changes in the distribution of different classes in task 1, and retains the knowledge of historical tasks with training.
This is mainly because our model learns through supervised comparison, and explicitly emphasizes that the samples in historical memory are compact within the class and far away from each other. And the knowledge of historical memory is preserved through the distillation of memory knowledge. Because knowledge distillation preserves the distance distribution between classes, it can make up for the contrastive learning to over-optimize the distance between classes to prevent overfitting.

\section{Conclusions and Future Work}
This paper proposes a novel consistent representation learning method for the CRE task, mainly through contrastive learning and knowledge distillation when replaying memory. Specifically, we use supervised contrastive learning based on a memory bank to train each new task so that the model can effectively learn the feature representation. In addition, in order to prevent the catastrophic forgetting of the old task, we conduct contrastive replay for memory samples, and at the same time, make the model retain the knowledge of the relation between the historical tasks through the knowledge distillation. Our method can better learn consistent representations to alleviate catastrophic forgetting effectively. Extensive experiments on two benchmark datasets show that our method significantly improves the performance of the most advanced technology and demonstrates powerful representation learning capabilities. In the future, we will continue to study cross-domain continual relation extraction to acquire ever-increasing knowledge.

\section*{Acknowledgments}
This paper is funded by National Natural Science Foundation of China (Grant No. 62173195) and Natural Science Foundation of Hebei Province, China (pre-research No. F2022208006). We would like to thank the anonymous reviewers for their valuable feedback.

\bibliography{mybib.bib, mybib2.bib, mybibfile.bib}

\begin{thebibliography}{34}
\expandafter\ifx\csname natexlab\endcsname\relax\def\natexlab#1{#1}\fi

\bibitem[{Aljundi et~al.(2018)Aljundi, Babiloni, Elhoseiny, Rohrbach, and
  Tuytelaars}]{aljundi2018memory}
Rahaf Aljundi, Francesca Babiloni, Mohamed Elhoseiny, Marcus Rohrbach, and
  Tinne Tuytelaars. 2018.
\newblock Memory aware synapses: Learning what (not) to forget.
\newblock In \emph{Proceedings of the European Conference on Computer Vision
  (ECCV)}, pages 139--154.

\bibitem[{Chaudhry et~al.(2018)Chaudhry, Ranzato, Rohrbach, and
  Elhoseiny}]{chaudhry2018efficient}
Arslan Chaudhry, Marc'Aurelio Ranzato, Marcus Rohrbach, and Mohamed Elhoseiny.
  2018.
\newblock Efficient lifelong learning with a-gem.
\newblock \emph{arXiv preprint arXiv:1812.00420}.

\bibitem[{Chen et~al.(2015)Chen, Goodfellow, and Shlens}]{chen2015net2net}
Tianqi Chen, Ian Goodfellow, and Jonathon Shlens. 2015.
\newblock Net2net: Accelerating learning via knowledge transfer.
\newblock \emph{arXiv preprint arXiv:1511.05641}.

\bibitem[{Chen et~al.(2020)Chen, Kornblith, Norouzi, and
  Hinton}]{chen2020simple}
Ting Chen, Simon Kornblith, Mohammad Norouzi, and Geoffrey Hinton. 2020.
\newblock A simple framework for contrastive learning of visual
  representations.
\newblock In \emph{International conference on machine learning}, pages
  1597--1607. PMLR.

\bibitem[{Chen and He(2021)}]{chen2021exploring}
Xinlei Chen and Kaiming He. 2021.
\newblock Exploring simple siamese representation learning.
\newblock In \emph{Proceedings of the IEEE/CVF Conference on Computer Vision
  and Pattern Recognition}, pages 15750--15758.

\bibitem[{Cui et~al.(2021)Cui, Yang, Yu, Hu, Cheng, Yi, and
  Xiao}]{cui2021refining}
Li~Cui, Deqing Yang, Jiaxin Yu, Chengwei Hu, Jiayang Cheng, Jingjie Yi, and
  Yanghua Xiao. 2021.
\newblock Refining sample embeddings with relation prototypes to enhance
  continual relation extraction.
\newblock In \emph{Proceedings of the 59th Annual Meeting of the Association
  for Computational Linguistics and the 11th International Joint Conference on
  Natural Language Processing (Volume 1: Long Papers)}, pages 232--243.

\bibitem[{d'Autume et~al.(2019)d'Autume, Ruder, Kong, and
  Yogatama}]{d2019episodic}
Cyprien de~Masson d'Autume, Sebastian Ruder, Lingpeng Kong, and Dani Yogatama.
  2019.
\newblock Episodic memory in lifelong language learning.
\newblock \emph{arXiv preprint arXiv:1906.01076}.

\bibitem[{Devlin et~al.(2019)Devlin, Chang, Lee, and
  Toutanova}]{DBLP:conf/naacl/DevlinCLT19}
Jacob Devlin, Ming{-}Wei Chang, Kenton Lee, and Kristina Toutanova. 2019.
\newblock \href {https://doi.org/10.18653/v1/n19-1423} {{BERT:} pre-training of
  deep bidirectional transformers for language understanding}.
\newblock In \emph{Proceedings of the 2019 Conference of the North American
  Chapter of the Association for Computational Linguistics: Human Language
  Technologies, {NAACL-HLT} 2019, Minneapolis, MN, USA, June 2-7, 2019, Volume
  1 (Long and Short Papers)}, pages 4171--4186. Association for Computational
  Linguistics.

\bibitem[{Dong et~al.(2021)Dong, Hong, Tao, Chang, Wei, and Gong}]{dong2021few}
Songlin Dong, Xiaopeng Hong, Xiaoyu Tao, Xinyuan Chang, Xing Wei, and Yihong
  Gong. 2021.
\newblock Few-shot class-incremental learning via relation knowledge
  distillation.
\newblock In \emph{Proceedings of the AAAI Conference on Artificial
  Intelligence}, volume~35, pages 1255--1263.

\bibitem[{Fernando et~al.(2017)Fernando, Banarse, Blundell, Zwols, Ha, Rusu,
  Pritzel, and Wierstra}]{fernando2017pathnet}
Chrisantha Fernando, Dylan Banarse, Charles Blundell, Yori Zwols, David Ha,
  Andrei~A Rusu, Alexander Pritzel, and Daan Wierstra. 2017.
\newblock Pathnet: Evolution channels gradient descent in super neural
  networks.
\newblock \emph{arXiv preprint arXiv:1701.08734}.

\bibitem[{Han et~al.(2020)Han, Dai, Gao, Lin, Liu, Li, Sun, and
  Zhou}]{han2020continual}
Xu~Han, Yi~Dai, Tianyu Gao, Yankai Lin, Zhiyuan Liu, Peng Li, Maosong Sun, and
  Jie Zhou. 2020.
\newblock Continual relation learning via episodic memory activation and
  reconsolidation.
\newblock In \emph{Proceedings of the 58th Annual Meeting of the Association
  for Computational Linguistics}, pages 6429--6440.

\bibitem[{Han et~al.(2018)Han, Zhu, Yu, Wang, Yao, Liu, and
  Sun}]{han-etal-2018-fewrel}
Xu~Han, Hao Zhu, Pengfei Yu, Ziyun Wang, Yuan Yao, Zhiyuan Liu, and Maosong
  Sun. 2018.
\newblock \href {https://doi.org/10.18653/v1/D18-1514} {{F}ew{R}el: A
  large-scale supervised few-shot relation classification dataset with
  state-of-the-art evaluation}.
\newblock In \emph{Proceedings of the 2018 Conference on Empirical Methods in
  Natural Language Processing}, pages 4803--4809, Brussels, Belgium.
  Association for Computational Linguistics.

\bibitem[{He et~al.(2020)He, Fan, Wu, Xie, and Girshick}]{he2020momentum}
Kaiming He, Haoqi Fan, Yuxin Wu, Saining Xie, and Ross Girshick. 2020.
\newblock Momentum contrast for unsupervised visual representation learning.
\newblock In \emph{Proceedings of the IEEE/CVF Conference on Computer Vision
  and Pattern Recognition}, pages 9729--9738.

\bibitem[{Hendrycks and Dietterich(2019)}]{hendrycks2019benchmarking}
Dan Hendrycks and Thomas Dietterich. 2019.
\newblock Benchmarking neural network robustness to common corruptions and
  perturbations.
\newblock \emph{arXiv preprint arXiv:1903.12261}.

\bibitem[{Jaiswal et~al.(2021)Jaiswal, Babu, Zadeh, Banerjee, and
  Makedon}]{jaiswal2021survey}
Ashish Jaiswal, Ashwin~Ramesh Babu, Mohammad~Zaki Zadeh, Debapriya Banerjee,
  and Fillia Makedon. 2021.
\newblock A survey on contrastive self-supervised learning.
\newblock \emph{Technologies}, 9(1):2.

\bibitem[{Khosla et~al.(2020)Khosla, Teterwak, Wang, Sarna, Tian, Isola,
  Maschinot, Liu, and Krishnan}]{khosla2020supervised}
Prannay Khosla, Piotr Teterwak, Chen Wang, Aaron Sarna, Yonglong Tian, Phillip
  Isola, Aaron Maschinot, Ce~Liu, and Dilip Krishnan. 2020.
\newblock Supervised contrastive learning.
\newblock \emph{arXiv preprint arXiv:2004.11362}.

\bibitem[{Kirkpatrick et~al.(2017)Kirkpatrick, Pascanu, Rabinowitz, Veness,
  Desjardins, Rusu, Milan, Quan, Ramalho, Grabska-Barwinska
  et~al.}]{kirkpatrick2017overcoming}
James Kirkpatrick, Razvan Pascanu, Neil Rabinowitz, Joel Veness, Guillaume
  Desjardins, Andrei~A Rusu, Kieran Milan, John Quan, Tiago Ramalho, Agnieszka
  Grabska-Barwinska, et~al. 2017.
\newblock Overcoming catastrophic forgetting in neural networks.
\newblock \emph{Proceedings of the national academy of sciences},
  114(13):3521--3526.

\bibitem[{Li et~al.(2020)Li, Zhou, Xiong, and Hoi}]{li2020prototypical}
Junnan Li, Pan Zhou, Caiming Xiong, and Steven~CH Hoi. 2020.
\newblock Prototypical contrastive learning of unsupervised representations.
\newblock \emph{arXiv preprint arXiv:2005.04966}.

\bibitem[{Liu and Abbeel(2020)}]{liu2020hybrid}
Hao Liu and Pieter Abbeel. 2020.
\newblock Hybrid discriminative-generative training via contrastive learning.
\newblock \emph{arXiv preprint arXiv:2007.09070}.

\bibitem[{Liu et~al.(2018)Liu, Masana, Herranz, Van~de Weijer, Lopez, and
  Bagdanov}]{liu2018rotate}
Xialei Liu, Marc Masana, Luis Herranz, Joost Van~de Weijer, Antonio~M Lopez,
  and Andrew~D Bagdanov. 2018.
\newblock Rotate your networks: Better weight consolidation and less
  catastrophic forgetting.
\newblock In \emph{2018 24th International Conference on Pattern Recognition
  (ICPR)}, pages 2262--2268. IEEE.

\bibitem[{Lopez-Paz and Ranzato(2017)}]{lopez2017gradient}
David Lopez-Paz and Marc'Aurelio Ranzato. 2017.
\newblock Gradient episodic memory for continual learning.
\newblock \emph{Advances in neural information processing systems},
  30:6467--6476.

\bibitem[{Mai et~al.(2021)Mai, Li, Kim, and Sanner}]{mai2021supervised}
Zheda Mai, Ruiwen Li, Hyunwoo Kim, and Scott Sanner. 2021.
\newblock Supervised contrastive replay: Revisiting the nearest class mean
  classifier in online class-incremental continual learning.
\newblock In \emph{Proceedings of the IEEE/CVF Conference on Computer Vision
  and Pattern Recognition}, pages 3589--3599.

\bibitem[{Soares et~al.(2019{\natexlab{a}})Soares, FitzGerald, Ling, and
  Kwiatkowski}]{soares2019matching}
Livio~Baldini Soares, Nicholas FitzGerald, Jeffrey Ling, and Tom Kwiatkowski.
  2019{\natexlab{a}}.
\newblock Matching the blanks: Distributional similarity for relation learning.
\newblock In \emph{Proceedings of ACL}.

\bibitem[{Soares et~al.(2019{\natexlab{b}})Soares, FitzGerald, Ling, and
  Kwiatkowski}]{DBLP:conf/acl/SoaresFLK19}
Livio~Baldini Soares, Nicholas FitzGerald, Jeffrey Ling, and Tom Kwiatkowski.
  2019{\natexlab{b}}.
\newblock \href {https://doi.org/10.18653/v1/p19-1279} {Matching the blanks:
  Distributional similarity for relation learning}.
\newblock In \emph{Proceedings of the 57th Conference of the Association for
  Computational Linguistics, {ACL} 2019, Florence, Italy, July 28- August 2,
  2019, Volume 1: Long Papers}, pages 2895--2905. Association for Computational
  Linguistics.

\bibitem[{Sun et~al.(2019)Sun, Ho, and Lee}]{sun2019lamol}
Fan-Keng Sun, Cheng-Hao Ho, and Hung-Yi Lee. 2019.
\newblock Lamol: Language modeling for lifelong language learning.
\newblock \emph{arXiv preprint arXiv:1909.03329}.

\bibitem[{Tao et~al.(2018)Tao, Gao, Shang, Wu, Zhao, and
  Yan}]{DBLP:conf/ijcai/TaoGSWZY18}
Chongyang Tao, Shen Gao, Mingyue Shang, Wei Wu, Dongyan Zhao, and Rui Yan.
  2018.
\newblock \href {https://doi.org/10.24963/ijcai.2018/614} {Get the point of my
  utterance! learning towards effective responses with multi-head attention
  mechanism}.
\newblock In \emph{Proceedings of the Twenty-Seventh International Joint
  Conference on Artificial Intelligence, {IJCAI} 2018, July 13-19, 2018,
  Stockholm, Sweden}, pages 4418--4424. ijcai.org.

\bibitem[{Wang et~al.(2019)Wang, Xiong, Yu, Guo, Chang, and
  Wang}]{wang2019sentence}
Hong Wang, Wenhan Xiong, Mo~Yu, Xiaoxiao Guo, Shiyu Chang, and William~Yang
  Wang. 2019.
\newblock Sentence embedding alignment for lifelong relation extraction.
\newblock \emph{arXiv preprint arXiv:1903.02588}.

\bibitem[{Wu et~al.(2021)Wu, Li, Li, Haffari, Qi, Zhu, and
  Xu}]{wu2021curriculum}
Tongtong Wu, Xuekai Li, Yuan-Fang Li, Reza Haffari, Guilin Qi, Yujin Zhu, and
  Guoqiang Xu. 2021.
\newblock Curriculum-meta learning for order-robust continual relation
  extraction.
\newblock \emph{CoRR, abs/2101.01926}.

\bibitem[{Wu et~al.(2018)Wu, Xiong, Yu, and Lin}]{wu2018unsupervised}
Zhirong Wu, Yuanjun Xiong, Stella~X Yu, and Dahua Lin. 2018.
\newblock Unsupervised feature learning via non-parametric instance
  discrimination.
\newblock In \emph{Proceedings of the IEEE conference on computer vision and
  pattern recognition}, pages 3733--3742.

\bibitem[{Xiong et~al.(2017)Xiong, Power, and Callan}]{DBLP:conf/www/XiongPC17}
Chenyan Xiong, Russell Power, and Jamie Callan. 2017.
\newblock \href {https://doi.org/10.1145/3038912.3052558} {Explicit semantic
  ranking for academic search via knowledge graph embedding}.
\newblock In \emph{Proceedings of the 26th International Conference on World
  Wide Web, {WWW} 2017, Perth, Australia, April 3-7, 2017}, pages 1271--1279.
  {ACM}.

\bibitem[{Yan et~al.(2021)Yan, Xie, and He}]{yan2021dynamically}
Shipeng Yan, Jiangwei Xie, and Xuming He. 2021.
\newblock Der: Dynamically expandable representation for class incremental
  learning.
\newblock In \emph{Proceedings of the IEEE/CVF Conference on Computer Vision
  and Pattern Recognition}, pages 3014--3023.

\bibitem[{Zenke et~al.(2017)Zenke, Poole, and Ganguli}]{zenke2017continual}
Friedemann Zenke, Ben Poole, and Surya Ganguli. 2017.
\newblock Continual learning through synaptic intelligence.
\newblock In \emph{International Conference on Machine Learning}, pages
  3987--3995. PMLR.

\bibitem[{Zhang et~al.(2017)Zhang, Zhong, Chen, Angeli, and
  Manning}]{zhang2017position}
Yuhao Zhang, Victor Zhong, Danqi Chen, Gabor Angeli, and Christopher~D Manning.
  2017.
\newblock Position-aware attention and supervised data improve slot filling.
\newblock In \emph{Proceedings of the 2017 Conference on Empirical Methods in
  Natural Language Processing}, pages 35--45.

\bibitem[{Zhou et~al.(2016)Zhou, Shi, Tian, Qi, Li, Hao, and
  Xu}]{zhou2016attention}
Peng Zhou, Wei Shi, Jun Tian, Zhenyu Qi, Bingchen Li, Hongwei Hao, and Bo~Xu.
  2016.
\newblock Attention-based bidirectional long short-term memory networks for
  relation classification.
\newblock In \emph{Proceedings of the 54th annual meeting of the association
  for computational linguistics (volume 2: Short papers)}, pages 207--212.

\end{thebibliography}
\bibliographystyle{acl_natbib}




\end{document}